\documentclass{article}

\PassOptionsToPackage{numbers, compress}{natbib}

\usepackage[main,final]{neurips_2026}

\usepackage{subcaption}

\usepackage[utf8]{inputenc} %
\usepackage[T1]{fontenc}    %
\usepackage{hyperref}       %
\usepackage{url}            %
\usepackage{booktabs}       %
\usepackage{amsfonts}       %
\usepackage{nicefrac}       %
\usepackage{microtype}      %
\usepackage{xcolor}         %
\usepackage{xspace}
\usepackage[frozencache]{minted}
\usepackage{soul}
\usepackage{tabularray}
\usepackage{graphicx}
\usepackage{enumitem}
\usepackage{wrapfig}

\definecolor{ForestGreen}{RGB}{34,139,34}
\newcommand{\highlightRelevantValue}[1]{%
  \ifdim #1pt > 0pt
    {\color{ForestGreen}{#1\%}}%
  \else
    {\color{red}{#1\%}}%
  \fi
}

\setlist{nosep}

\UseTblrLibrary{booktabs}
\newmintinline[code]{text}{}

\DeclareRobustCommand{\projectname}{\textsc{WebServ}\xspace}

\title{
\projectname: A Full-Stack and RL-Ready Web Environment for Training Web Agents at Scale 
}

\author{
Yuxuan Lu\thanks{Work was done while interning at Amazon.} \\
Northeastern University \\
\texttt{lu.yuxuan@northeastern.edu}
\And
Ziyi Wang\footnotemark[1] \\
Northeastern University
\And
Jing Huang \\
Amazon
\And
Hui Liu \\
Amazon
\And
Jiri Gesi \\
Amazon
\And
Yan Han \\
Amazon
\And
Shihan Fu\footnotemark[1] \\
Northeastern University
\And
Tianqi Zheng \\
Amazon
\And
Xianfeng Tang \\
Amazon
\And
Chen Luo \\
Amazon
\And
Yisi Sang \\
Amazon
\And
Jin Lai \\
Amazon
\And
Dakuo Wang\thanks{Work was done while visiting Amazon.} \\
Northeastern University \\
\texttt{d.wang@northeastern.edu}
}

\begin{document}

\maketitle

\begin{abstract}
Reinforcement learning (RL) for web agents demands environments that are both effective for evaluation and efficient enough for large-scale on-policy training. Current web environments fall short: server-side Docker setups are too resource-intensive for massive parallel rollouts, while browser-side interfaces produce noisy observations, execute actions unreliably under modern single-page applications, and omit visual interactivity cues. We introduce \projectname, a full-stack, RL-ready web environment that addresses these limitations end-to-end. On the server side, \projectname uses \textbf{Incus containers} with block-level copy-on-write, reducing launch latency by \textbf{$\sim$5$\times$} and persistent storage by \textbf{$\sim$240$\times$}, enabling \textbf{200+} concurrent isolated environments on a single host. On the browser side, \projectname provides a compact, site-agnostic \textbf{observation and action interface} derived automatically from the DOM with human-aligned interactivity cues, and a robust \textbf{action execution backend} using network-aware waiting for reliable SPA support. On WebArena-Lite, \projectname achieves state-of-the-art single-prompt results, with controlled comparisons confirming consistent gains across GPT-4o, OpenAI-o3, and Llama-3.1-8B over vanilla WebArena. We further train Qwen3-4B and Qwen3-30B-A3B with RL entirely within \projectname; the RL-trained 4B model achieves \textbf{57.3\%} mean accuracy, surpassing both Claude~4.5 Sonnet (50.0\%) and the RL-trained 8B model from WebAgent-R1 (51.8\%).
\end{abstract}

\section{Introduction}

Web Agents that perceive and act on real user interfaces are increasingly applied to complex, multi-step tasks such as automated UI testing \cite{luUXAgentLLMAgentBased2025}, question answering \cite{nakanoWebGPTBrowserassistedQuestionanswering2022}, and privacy-preserving browsing \cite{chenEmpathyBasedSandboxApproach2024}.
While early approaches relied on prompting \cite{sodhiStePStackedLLM2024,maLASERLLMAgent2024} or supervised fine-tuning \cite{luPromptingNotAll2025}, recent work has turned to Reinforcement Learning (RL) to enable self-improvement through interaction \cite{weiWebAgentR1TrainingWeb2025,qiWebRLTrainingLLM2025,deepseek-aiDeepSeekR1IncentivizingReasoning2025}.
However, progress in RL-based Web Agents remains bottlenecked not by algorithms, but by the absence of a full-stack web environment that is both effective for evaluation and efficient enough for large-scale on-policy training.

\begin{wrapfigure}{r}{0.5\textwidth}
  \centering
  \includegraphics[width=0.5\textwidth]{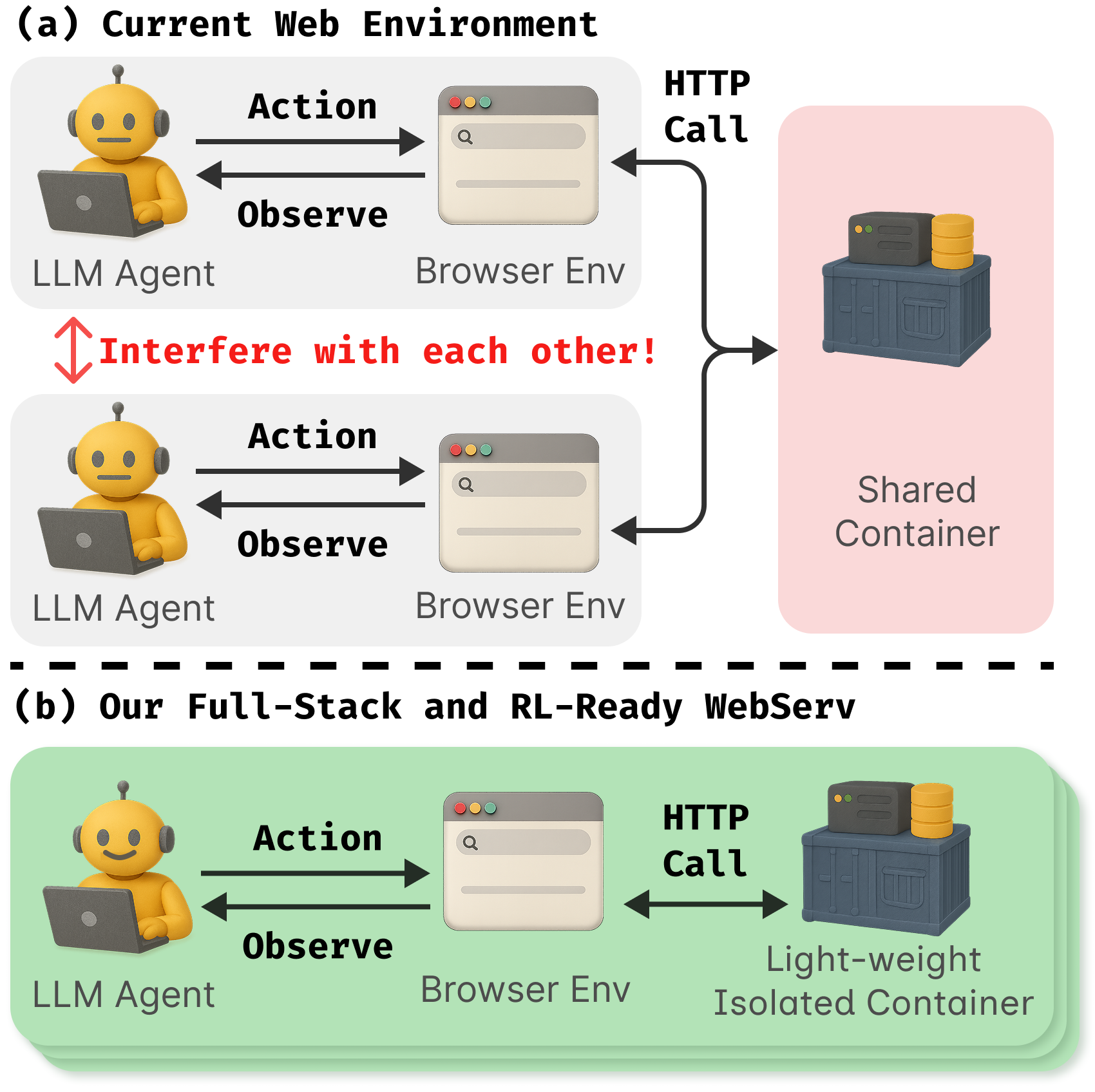}
  \caption{System architecture of \projectname. (a) existing web environments rely on a shared server container that leads to cross agent interference; (b) \projectname creates lightweight per agent containers to ensure isolation and scalable rollouts for RL training.}
  \label{fig:teaser}
\end{wrapfigure}
Researchers have concluded that the current generation of web agent environments is ``not RL-ready'' \cite{chenScalingAgentLearning2025}. 
On the \textbf{server side}, existing environments such as WebArena \cite{zhouWebArenaRealisticWeb2024} use Docker containers for reproducibility; however, these containers are resource-intensive (launching a single shopping environment requires approximately 6\,GB of storage and about one minute to start), which \textbf{limits large-scale parallel evaluation and massive RL rollouts}.
On the \textbf{browser side}, current environments suffer from multiple issues.
Agents struggle to generalize across websites because many environments require manual, per-site observation engineering or depend on site-specific features such as accessibility trees, which are often missing or inconsistent \cite{zhouWebArenaRealisticWeb2024,yaoWebShopScalableRealWorld2022}.
Context windows are overwhelmed and action selection becomes unnecessarily difficult when systems pass raw HTML as observations and allow arbitrary Python code as actions \cite{gurRealWorldWebAgentPlanning2023}.

In addition, actions may fail silently or produce stale observations because many systems wait a fixed duration before returning the next state, which breaks under modern Single-Page Applications (SPAs) where asynchronous updates arrive at unpredictable times.
Moreover, models frequently click non-interactive elements because most setups omit visual cues \cite{zhouWebArenaRealisticWeb2024,lutzWILBURAdaptiveInContext2024} that humans rely on to judge interactivity (e.g., pointer cursor on clickable elements, text cursor on inputs) \cite{luPromptingNotAll2025}. As a result, models waste steps recovering from avoidable interaction errors.
Taken together, these server-side and browser-side limitations impede progress toward scalable and robust Web Agents.

We introduce \projectname, a full-stack and RL-ready web environment for scalable RL web-agent training and evaluation (Fig. \ref{fig:teaser}).
\projectname addresses the above issues through coordinated improvements across the browser and server stack. On the browser side, it provides a compact, site-agnostic observation and action interface that is derived automatically from the web page. The interface preserves human-aligned interactivity cues so that agents can identify actionable elements. \projectname also provides a robust action execution backend that synchronizes state transitions using network-aware waiting. This design improves reliability for SPA interactions and reduces variability across repeated runs. On the web server side, \projectname uses Incus containers with block-level copy-on-write. This mechanism supports fast launching, and cloning any existing Docker-based web applications. This method makes it practical to run hundreds of isolated client–server rollouts on a single host.

To measure the effectiveness of our browser I/O interface design, we evaluate prompting-based agents on WebArena-Lite, where \projectname paired with Claude~4.5 achieves a \textbf{40.0\%} success rate on the Shopping task, \textbf{62.9\%} on the CMS task, and \textbf{50.0\%} on the GitLab task, establishing a new state-of-the-art among single-prompt agents. Controlled comparisons confirm that the same models (GPT-4o, OpenAI-o3, Llama-3.1-8B) consistently improve when switching from vanilla WebArena to \projectname, isolating the gains to our environment design.
On the server side, \projectname reduces launch latency by \textbf{$\sim$5$\times$} and persistent storage by \textbf{$\sim$240$\times$}, enabling \textbf{200+} concurrent containers on a single host.
To validate that these efficiency gains translate into effective RL training, we train Qwen3-4B and Qwen3-30B-A3B with on-policy reinforcement learning entirely within \projectname. The RL-trained 4B model achieves 57.3\% mean accuracy, surpassing both Claude~4.5 Sonnet (50.0\%) and the RL-trained Llama-3.1-8B from WebAgent-R1 (51.8\%) -- demonstrating that a small open-source model trained with RL in our environment can outperform frontier proprietary models and existing RL-trained agents. The 30B model reaches 61.8\% mean accuracy.

To summarize, our contributions are as follows:
\begin{itemize}[leftmargin=*]
\item \textbf{Efficient and effective browser I/O interface} A site-agnostic interface that shortens the context by removing invisible or noisy elements and exposes a small, well-defined set of actions with simple parameters, while maintaining generalizability to real websites or Dockerized web applications.
\item \textbf{Robust action execution.} A robust action executor that locates the target element, performs the action and returns the next observation only after the page has finished updating (for example, after network requests finish). The executor uses limited retries and clear error messages, and generalizes across sites.
\item \textbf{Scalable environment for RL.} An \textit{Incus} based manager that can start, clone, and reset a paired browser and web server quickly (sub-second startup), so many runs can proceed in parallel with low resource use. We integrate our environment as tools for VeRL and Slime\footnote{Code available at \url{https://anonymous.4open.science/r/webserv_anonymous-90EF/}} for out-of-the-box usage. We validate this capability by training Qwen3-4B and Qwen3-30B-A3B with RL, where the 4B model surpasses both Claude~4.5 Sonnet and the RL-trained 8B model from WebAgent-R1.
\end{itemize}

\section{Related Works}

\subsection{Web Environment Design}
A key challenge in designing browser-based environments is balancing between the complexity of the observation space and the flexibility of the action space. 

For instance, WebShop \cite{yaoWebShopScalableRealWorld2022} introduces a simulated shopping environment where the observation space is task-specific. It consists of the current page (homepage, search results page, or product detail page) along with structured information such as the list of products. 
The action space is similarly constrained to a set of task-specific operations (e.g., search, purchase). 
While this design keeps the action space minimal yet semantically meaningful, it suffers from two major limitations: 
(1) the model requires a long and complex prompt to encode both the observation and action space definitions, and 
(2) the environment lacks generalizability across tasks and domains.

In contrast, WebAgent \cite{gurRealWorldWebAgentPlanning2023} represents the observation space using raw HTML and defines the action space as Python code that directly operates the browser. 
This design removes the need for manually specifying the action space and allows the model to leverage its inherent understanding of HTML elements (for example, recognizing an input field \code{<input type="text"/>} without requiring explicit prompt engineering). 
However, this approach introduces new challenges: the model must generate arbitrary code as actions, which requires not only familiarity with external libraries (such as Playwright) but also the ability to produce syntactically correct and semantically valid code.

To strike a middle ground between raw HTML and task-specific environments, WebArena \cite{zhouWebArenaRealisticWeb2024} introduces an accessibility-tree representation that extracts only the information necessary for completing web tasks, thereby reducing the noise of raw HTML. 
It also defines a custom action space consisting of operations such as clicks and inputs. 
This approach maintains a compact context while avoiding an overly complex action space. 
However, it still requires website-specific adaptations (e.g., accessibility support), limiting its generalizability across domains. 

From an implementation perspective, many existing frameworks re-parse the webpage after a page-load event following each action. 
This assumption often fails in real-world websites, where only parts of the page are refreshed, leading to incomplete or outdated observations. 
In addition, human users rely heavily on visual cues while browsing the web, such as cursor styles, to infer interactivity. 
For example, humans can easily distinguish between selectable text (indicated by ``\raisebox{-0.2ex}{\includegraphics[height=1.4ex]{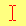}}'') and clickable elements (indicated by ``\raisebox{-0.2ex}{\includegraphics[height=1.4ex]{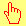}}''), whereas agents frequently attempt to click plain text or other non-interactive elements. 
Such discrepancies highlight the need for web environments that: 
(1) balance the complexity of context and action spaces, 
(2) ensure robust action execution on dynamic, real-world websites with complex network conditions, and 
(3) provide essential visual cues to better align agent perception with human browsing behavior.

\subsection{Web Agents}
Recent advances in LLM-based Web Agents have significantly enhanced the ability of LLMs to assist with web-based tasks.
Early work like WebGPT~\cite{nakanoWebGPTBrowserassistedQuestionanswering2022} enabled GPT models to interact with search engines, significantly improving question-answering performance. Subsequent systems, including WebVoyager~\cite{heWebVoyagerBuildingEndtoEnd2024}, LASER~\cite{maLASERLLMAgent2024}, and WebAgent~\cite{gurRealWorldWebAgentPlanning2023}, extended LLM capabilities to multimodal interaction, complex state space navigation, and long-horizon planning. Claude's Computer-use API\footnote{\url{https://www.anthropic.com/news/3-5-models-and-computer-use}} further demonstrated precise, general-purpose control over user interfaces beyond web browsers.

\subsection{RL Training of Web Agents}
Building on recent progress in improving the reasoning capabilities of LLMs through RL~\cite{deepseek-aiDeepSeekR1IncentivizingReasoning2025}, researchers have started to explore RL in web-based environments.
Web-RL~\cite{qiWebRLTrainingLLM2025} proposes a self-evolving curriculum learning framework in which agents generate new tasks from previous rollout trajectories. However, due to the low efficiency of on-policy algorithms, it relies on pre-generated off-policy rollouts.
WebAgent-R1~\cite{weiWebAgentR1TrainingWeb2025} introduces an online RL training paradigm for Web Agents, but its rollout setup allows multiple agents to interact with a shared web server instance, which can lead to interference and unstable training.
More recently, \citet{chenScalingAgentLearning2025} argue that web tasks are not RL Ready and propose an environment synthesis approach to generate RL training data at scale.
Together, these works highlight a growing recognition in the RL community of the need for scalable and efficient environment to manage server containers and support large-scale web agent training.

\section{\projectname}

\subsection{System Overview}
To address challenges in existing Web Environments, we propose \projectname, a full-stack framework that integrates: 1) a compact, site-agnostic \textbf{observation and action interface} derived automatically from the DOM, and it preserves human-aligned interactivity cues; 2) a robust \textbf{action execution backend} that synchronizes state transitions using network-aware quiescence for reliable SPA interactions; and 3) an Incus-based container manager that supports fast launching, cloning, and rollback of full web applications.

\subsection{Environment}

\subsubsection{Observation Space}
A central challenge identified in prior work is that raw HTML observations overwhelm models with excessive and noisy content, while handcrafted task-specific abstractions (for example, WebShop) lack generality across domains.

To address this, \projectname employs a DOM parser that automatically reduces the page to elements that are visible and meaningful to human users. The parser filters invisible and non-semantic elements, flattens redundant container nesting, detects interactive elements via heuristics (native controls, ARIA roles, cursor styles), and assigns each interactive element a stable, human-readable semantic identifier. The final observation is a compact JSON object containing stripped HTML, lists of clickable/hoverable elements, and input/select state. Full details of the observation construction pipeline are provided in Appendix~\ref{sec:observation-space}.

\subsubsection{Visual Observation Space}
Beyond text-only agents, recent work has begun to explore Visual Language Models (VLMs \cite{yinSurveyMultimodalLarge2024}) in web-based environments \cite{kohVisualWebArenaEvaluatingMultimodal2024}. To support these use cases, \projectname provides a VLM mode in which a screenshot of the currently observed webpage is captured and provided to the model as part of the observation.

\subsubsection{Action Space}
Another challenge is that prior systems define action spaces at extremes: WebAgent relies on arbitrary code generation in Python, which requires library-specific knowledge and produces unstable executions, while WebShop restricts the agent to narrow, task-specific actions that do not generalize.  

We design the action space so that agents can perform the same primitive interactions as human users (click, type, hover, select, navigate, manage tabs). Full action space can be found in Appendix \ref{sec:action-space}. To reduce latency and eliminate unnecessary branching, we explicitly remove \emph{scroll} action (the environment automatically scrolls the target element into view before execution). Targets are referenced by the stable semantic identifiers described in the observation construction (\code{data-semantic-id}), which makes actions robust to minor DOM changes. 
Before execution, the environment validates that the target remains interactable, scrolls it into view automatically (therefore no explicit scroll action is necessary), performs the operation, and then waits for an idle period before emitting the next observation. 
After each action, the executor waits for network and UI quiescence (e.g., no active requests during a short idle window) before returning the next observation. This design addresses prior challenges of noisy action choices, arbitrary code generation, and nondeterministic post-action states by keeping the action set compact, interpretable, and synchronized with the rendered page.

\subsection{Robust Action Execution}

Prior environments typically enforce a fixed waiting time after each action or assume that a new page will trigger a complete reload event. However, static sleep intervals either waste time or fail to capture late-arriving updates, while relying on full page loads is incompatible with modern single-page applications (SPAs) where only parts of the DOM refresh in response to actions. As a result, agents may observe incomplete or inconsistent states, leading to unstable behavior.

To address this issue, \projectname introduces a network-aware synchronization mechanism that hooks into the browser's JavaScript runtime. We intercept both \code{XMLHttpRequest} and \code{fetch} APIs to track active requests and update a global activity counter whenever a request is initiated or completed. This instrumentation enables us to detect fine-grained network activity across the page, regardless of whether the site uses traditional reloads or partial updates.

After each agent action, the environment delays the next observation until it has observed a configurable idle period (e.g., 500\,ms) with no outstanding network requests. This guarantees that dynamically loaded content (such as search results, shopping carts, or dropdowns populated asynchronously) is fully rendered before being exposed to the agent. If the page fails to reach an idle state within a timeout window, the environment returns control with an explicit error state, allowing the agent or training loop to handle the failure deterministically.

By synchronizing observations with actual network quiescence rather than fixed delays or reload events, \projectname provides robust action execution on real-world websites that depend heavily on asynchronous content loading.

\subsection{Scalable and Efficient Web Server Manager}

\begin{table*}[t]
\centering
\caption{Comparison of model accuracy (in \%) across Shopping, CMS and GitLab tasks in WebArena-Lite \cite{qiWebRLTrainingLLM2025}. Single-prompt models, when paired with WebServ, outperform multi-prompt and reasoning-based baselines and achieve performance comparable to, or exceeding, that of RL-trained models.}
\begin{booktabs}{
  colspec = {X c c c c},
  width=\linewidth,
  row{1} = {font=\bfseries},
  cell{2,7}{1} = {c=5}{bg=black!30!white,font=\bfseries},
}
\toprule
Model and Method & Type & Shopping & CMS & GitLab \\
\midrule
Vanilla WebArena \\
GPT-4o \cite{weiWebAgentR1TrainingWeb2025} & Single-Prompt & 11.1 & 20 & 10.0 \\
OpenAI-o3 \cite{weiWebAgentR1TrainingWeb2025} & Reasoning Model & 33.3 & 45.7 & 46.7 \\
Llama-3.1-8B \cite{weiWebAgentR1TrainingWeb2025} & Single-Prompt & 8.9 & 5.7 & 10.0 \\
Llama3.1-8B + WebAgent-R1 \cite{weiWebAgentR1TrainingWeb2025} & Reinforcement Learning & \textbf{44.4} & \textbf{57.1} & \textbf{56.7} \\
\projectname \\
GPT-4o  & Single-Prompt  & 20.0 & 28.6 & 43.3 \\
OpenAI-o3  & Reasoning Model  & 37.8 & 48.6 & 46.7 \\
Llama-3.1-8B  & Single-Prompt  & 11.1 & 11.4 & 16.7 \\
Qwen3-4B  & Single-Prompt  & 22.2 & 22.9 & 30.0 \\
Qwen3-30B-A3B  & Single-Prompt  & 26.7 & 40.0 & 43.3 \\
GPT-5  & Single-Prompt & 35.6 & 57.1 & \textbf{53.3} \\
Claude 3.5 Sonnet  & Single-Prompt  & 26.7 & 31.4 & 36.7 \\
Claude 3.7 Sonnet  & Single-Prompt  & 31.1 & 37.1 & 50.0 \\
Claude 4 Sonnet  & Single-Prompt  & \textbf{42.2} & 48.6 & 50.0 \\
Claude 4.5 Sonnet  & Single-Prompt  & 40.0 & \textbf{62.9} & 50.0 \\
\bottomrule
\end{booktabs}
\label{tab:model_results}
\vspace{2\baselineskip}
\end{table*}

To enhance reproducibility, recent works such as WebArena \cite{zhouWebArenaRealisticWeb2024} distribute environments as Docker images. However, Docker is designed for long-lived services, not the repeated resets required by RL: launching a single WebArena shopping container requires several gigabytes of storage and tens of seconds of startup time, making large-scale rollouts infeasible.

We implement a scalable web server manager based on \emph{Incus}, a modern container runtime that extends Linux Containers (LXC). Incus uses block-level copy-on-write (via ZFS/Btrfs), so only modified blocks are duplicated on each reset—unlike Docker's file-level overlay, which copies entire multi-gigabyte files. This yields sub-second container startup and minimal storage overhead across parallel environments. Incus also retains OCI compatibility, allowing existing Docker-based web environments to run unmodified, and supports cloning of running containers for checkpointing and rollback during RL exploration. Together, these capabilities enable \projectname to scale to thousands of parallel rollouts on a single host.

\subsection{Post-Training Transfer and Human-Centered Inspection}

Although \projectname is designed primarily for scalable training and evaluation, we also consider the downstream transfer of trained agents to real-world browsing contexts. To bridge the gap between controlled environments and deployment settings, we develop human-centered inspection tools that make model behavior transparent and verifiable.

Together, these tools enable trained agents to move beyond offline simulation and be examined, audited, and adapted for practical integration. They also facilitate the collection of additional user feedback and correction signals that can guide further fine-tuning or reinforcement learning updates.

\section{Evaluation}

\begin{table}[t]
\centering
\caption{Comparison of system efficiency between \projectname and Docker.}
\begin{booktabs}{
  colspec = {l c c},
  row{1} = {font=\bfseries}
}
\toprule
Metric & \projectname (Incus) & Naïve Docker \\
\midrule
Launch speed & 1.781\,s & 8.963\,s \\
Storage & 28.01\,MiB & 6.78\,GiB \\
Memory & 1.74\,GiB & 1.63\,GiB \\
\bottomrule
\end{booktabs}
\label{tab:system_speed}
\end{table}

\subsection{Performance Benchmark on WebArena}

We evaluate \projectname on the Shopping, CMS, and GitLab tasks from WebArena-Lite \cite{qiWebRLTrainingLLM2025}, a 110-task subset of WebArena \cite{zhouWebArenaRealisticWeb2024} with human-verified evaluation rubrics. We focus on these three domains because they involve write actions that modify server-side state, which is where container isolation and fast resets are critical. (WebArena's Map and Wikipedia tasks are read-only and do not benefit from isolation; \projectname supports them but we omit them for this reason.) To compare with prior work \cite{weiWebAgentR1TrainingWeb2025}, we follow the same evaluation protocol.

Table~\ref{tab:model_results} summarizes results. Under a strict single-prompt setting, \projectname achieves competitive or superior performance to multi-prompt and RL-trained baselines, demonstrating that a clean, semantically enriched observation space alone enables LLMs to solve complex web tasks without multi-round reasoning or task-specific fine-tuning.

In addition, when comparing the performance of the same models on \projectname and on vanilla WebArena, \projectname consistently yields significant performance improvements across GPT-4o, OpenAI-o3, and Llama-3.1-8B, which further demonstrates that the observation and action space design of \projectname can substantially boost agent performance across different model categories (proprietary, reasoning, and open-source).

\begin{table*}[t]
\centering
\caption{Ablation study on the impact of removing visual cues on accuracy (\%) across sites. Each model is evaluated under the default setting and a variant with visual cues removed, highlighting the impact of visual information on task performance.}
\small
\begin{booktabs}{
colspec={Xccccccc},
column{1}={l},
cell{3,5,7,9,11,13,15,17,19}{1}={r,font=\itshape},
row{1}={font=\bfseries,c},
width=\linewidth
}
\toprule
Model & Shopping & CMS & GitLab & $\Delta$ Shopping (\%) & $\Delta$ CMS (\%) & $\Delta$ GitLab (\%) \\
\midrule
GPT-4o & 20.0 & 28.6 & 43.3 &  &  &  \\
w/o Visual Cue & 13.3 & 8.6 & 23.3 & \highlightRelevantValue{-33.3} & \highlightRelevantValue{-70.0} & \highlightRelevantValue{-46.2} \\
GPT-5 & 35.6 & 57.1 & 53.3 &  &  &  \\
w/o Visual Cue & 28.9 & 37.1 & 40.0 & \highlightRelevantValue{-18.8} & \highlightRelevantValue{-35.0} & \highlightRelevantValue{-25.0} \\
Claude 3.5 Sonnet & 26.7 & 31.4 & 36.7 &  &  &  \\
w/o Visual Cue & 11.1 & 17.1 & 6.7 & \highlightRelevantValue{-58.3} & \highlightRelevantValue{-45.5} & \highlightRelevantValue{-81.8} \\
Claude 3.7 Sonnet & 31.1 & 37.1 & 50.0 &  &  &  \\
w/o Visual Cue & 15.6 & 17.1 & 16.7 & \highlightRelevantValue{-50.0} & \highlightRelevantValue{-53.8} & \highlightRelevantValue{-66.7} \\
Claude 4 Sonnet & 42.2 & 48.6 & 50.0 &  &  &  \\
w/o Visual Cue & 31.1 & 34.3 & 33.3 & \highlightRelevantValue{-26.3} & \highlightRelevantValue{-29.4} & \highlightRelevantValue{-33.3} \\
Claude 4.5 Sonnet & 40.0 & 62.9 & 50.0 &  &  &  \\
w/o Visual Cue & 37.8 & 34.3 & 53.3 & \highlightRelevantValue{-5.6} & \highlightRelevantValue{-45.5} & \highlightRelevantValue{6.7} \\
\bottomrule
\end{booktabs}
\label{tab:visual_ablation}
\vspace{\baselineskip}
\end{table*}

\begin{table}[t]
\centering
\caption{Ablation results comparing text-only and VLM settings for Claude~4 Sonnet and Claude~4.5 Sonnet across WebArena-Lite tasks.}
\begin{booktabs}{
colspec={lccc},
cell{3,5}{1}={r}
}
\toprule
Model & Shopping & CMS & GitLab \\
\midrule
Claude 4 Sonnet & 42.2 & \textbf{48.6} & \textbf{50.0} \\
 + VLM & \textbf{44.4} & \textbf{48.6} & 43.3 \\
Claude 4.5 Sonnet & \textbf{40.0} & 62.9 & 50.0 \\
 + VLM & 37.8 & \textbf{65.7} & \textbf{56.7} \\
\bottomrule
\end{booktabs}

\vspace{\baselineskip}
\label{tab:vlm}
\end{table}

\subsection{System Speed and Resource Usage}

We benchmark \projectname's Incus-based container manager against a naïve Docker setup on an AWS EC2 \code{r6id.metal} instance (128 vCPUs, 1024\,GiB memory). Table~\ref{tab:system_speed} reports results for the WebArena shopping environment.

\projectname achieves an average launch time of $1.78$s (vs.\ $\sim$9s for Docker) and reduces storage need from $6.78$\,GiB to $28$\,MiB through block-level copy-on-write. Memory usage remains comparable ($1.74$\,GiB vs.\ $1.63$\,GiB), with the slight increase due to per-container namespace isolation. The primary scalability bottleneck is disk I/O: Docker writes $\sim$6\,GiB per launch, while \projectname writes only $\sim$28\,MiB, with reads heavily cached via ZFS. This enables 200+ concurrent containers on a single host with near-constant launch latency under parallel initialization.

\subsection{Reinforcement Learning Training}

\subsubsection{RL Training Efficiency and Speed}

To validate the scalability of \projectname for RL training, we measure end-to-end training throughput on the Qwen3-4B model. We set the maximum number of parallel rollout instances to 512, distributed across a small cluster of 4 AWS \code{r6id.16xlarge} CPU instances, each hosting Incus containers running the full web server stack. Training is distributed across 64 NVIDIA H200 GPUs (4 actor nodes and 4 rollout nodes, each with 8 GPUs), with 32 rollout workers serving on-policy generation. 

Under this configuration, \textbf{each RL step, comprising environment rollout, on-policy generation, reward computation, and gradient update, completes in approximately 12 minutes, yielding roughly 5 training steps per hour}. Each step collects 200 rollouts with multi-turn interactions averaging 11 turns and 77k tokens per trajectory. For reference, AReaL\cite{} reports approximately 6 minutes per step for a 7B model on 24 nodes (192 GPUs, 3$\times$ our GPU compute) on single-turn mathematical reasoning tasks. In contrast, \textbf{\projectname sustains comparable per-step latency with fewer GPUs on a substantially more complex workload}: multi-step web agent tasks that involve real network round-trips and dynamic server-side state. We can effectively feed large-scale GPU clusters with a very small CPU cluster, demonstrating that \projectname makes RL training on real web environments both practical and efficient.

\subsubsection{RL Agent Performance}

To validate that \projectname's scalable server management enables effective RL training, we train two open-source models (Qwen3-4B-thinking-2507 and Qwen3-30B-A3B-thinking-2507) using reinforcement learning with on-policy rollouts generated within our environment. We use GRPO \cite{shaoDeepSeekMathPushingLimits2024} with dynamic filtering \cite{yuDAPOOpenSourceLLM2025} as the RL algorithm. For both models, we first perform supervised fine-tuning bootstrap (SFT) with training trajectories generated from Claude 4.5 Sonnet for 3 epochs then employ RL with \projectname to further fine-tune performance. We report results at RL step~99 for both models. Full training hyperparameters are provided in Appendix~\ref{sec:rl-hyperparams}.

\begin{table}[t]
\centering
\caption{RL training results on WebArena-Lite. Base Model denotes the pre-trained model before any fine-tuning. SFT denotes the supervised fine-tuning checkpoint used to initialize RL. RL denotes the checkpoint at step~99. Accuracy is reported in \%.}
\begin{booktabs}{
colspec={llcccc},
row{1}={font=\bfseries},
}
\toprule
Model & Method & Shopping & CMS & GitLab & Mean \\
\midrule
Qwen3-4B & Single-Prompt & 22.2 & 22.9 & 30.0 & 24.5 \\
Qwen3-30B-A3B & Single-Prompt & 26.7 & 40.0 & 43.3 & 35.5 \\
Llama-3.1-8B + WebAgent-R1 & RL & 44.4 & 57.1 & 56.7 & 51.8 \\
Claude 4.5 Sonnet & Single-Prompt & 40.0 & 62.9 & 50.0 & 50.0 \\
\midrule
Qwen3-4B & SFT & 56.7 & 51.4 & 53.3 & 54.1 \\
Qwen3-4B & RL & 60.0 & 54.3 & 56.7 & \textbf{57.3} \\
Qwen3-30B-A3B & SFT & 64.4 & 54.3 & 53.3 & 58.2 \\
Qwen3-30B-A3B & RL & \textbf{71.1} & \textbf{57.1} & 53.3 & \textbf{61.8} \\
\bottomrule
\end{booktabs}
\label{tab:rl_results}
\end{table}

Table~\ref{tab:rl_results} shows that RL training within \projectname yields substantial improvements over the SFT baseline. The Qwen3-4B model improves from 54.1\% to 57.3\% mean accuracy after RL, surpassing both Claude~4.5 Sonnet (50.0\% mean) and the RL-trained Llama-3.1-8B from WebAgent-R1 (51.8\% mean). The Qwen3-30B-A3B model improves from 58.2\% to 61.8\% mean accuracy. These results demonstrate that \projectname's scalable container management and fast environment resets genuinely enable effective RL training for web agents.

\subsection{Ablations}

\subsubsection{Visual LM}

To study the contribution of visual signals in web agent tasks, we evaluate Claude~4 Sonnet and Claude~4.5 Sonnet under a VLM setting. In this configuration, a full page screenshot of the currently observed webpage is captured and provided to the model as an additional input, alongside the existing text-based HTML observation. This design allows the model to access visual cues such as layout, relative positioning, and visual grouping, which are absent from text-only representations.

Table~\ref{tab:vlm} reports the results. Overall, the impact of visual input varies across both models and task categories.
Taken together, these results suggest that visual signals are not universally beneficial for general web agent tasks.
Their effectiveness depends on both the underlying model capabilities and the nature of the task.

\subsubsection{Visual Cue}

A key feature of \projectname is its explicit preservation of visual cues that humans naturally rely on when interacting with webpages. In particular, \projectname identifies clickable elements in a human-aligned manner and exposes this information to the model by annotating interactive elements with the HTML attribute (\code{parser-clickable=true}), while assigning element IDs only to those elements. 
In contrast, prior environments such as WebArena~\cite{zhouWebArenaRealisticWeb2024} assign element IDs to all DOM nodes (or accessibility tree nodes) without distinguishing interactive from non-interactive elements. As a result, models may attempt to click on visually salient but non-actionable elements, leading to invalid actions and degraded performance.

To quantify the impact of these visual cues, we conduct an ablation study in which this information is removed. Specifically, we assign IDs to all elements and remove the \code{parser-clickable=true} annotation from the observation space, making interactive and non-interactive elements indistinguishable to the model. The results of this ablation are reported in Table~\ref{tab:visual_ablation}. 

Nearly all models exhibit a performance drop across almost all task settings, with degradations ranging from approximately 5\% to over 80\%, with the sole exception of Claude~4.5 Sonnet on the GitLab task. Models with lower baseline performance, such as Qwen~2.5~32B, GPT~4o~Mini, and Claude~3.5 Sonnet, experience substantially larger performance drops, indicating that explicit visual cues play a particularly important role for weaker or smaller models. 

Stronger models are better able to infer interactivity from context and surrounding text even without explicit cues, but they still generally show performance degradation. These results highlight the importance of explicitly encoding human-aligned visual signals in the observation space, both to stabilize agent behavior and to improve overall performance, especially in low-capacity models.

\section{Conclusion}

We introduce \projectname, a full-stack and RL-ready web environment for scalable web-agent training and evaluation. \projectname provides a compact DOM-derived observation and action interface, a robust action execution backend based on network-aware quiescence for modern SPAs, and an Incus-based server manager with block-level copy-on-write for fast isolation and reset. Across Shopping, CMS, and GitLab tasks in WebArena-Lite, \projectname enables state-of-the-art single-prompt performance and supports large-scale experimentation. It reduces container launch latency by approximately \textbf{$\sim$5$\times$}, reduces persistent storage by approximately \textbf{$\sim$240$\times$}, and supports launching and running \textbf{200+} concurrent containers on a single host within minutes. We further validate \projectname as an RL training platform: a 4B-parameter model trained with RL achieves 57.3\% mean accuracy, surpassing both Claude~4.5 Sonnet (50.0\%) and the RL-trained 8B model from WebAgent-R1 (51.8\%), and a 30B model reaches 61.8\% mean accuracy.

\section*{Limitations}

While \projectname provides a scalable and efficient environment for training and evaluating RL-based Web Agents, it has several limitations that open avenues for future work.

First, in text-only mode, although our parser produces a compact and semantically enriched HTML observation, as with all text-only environments, it does not retain visual \textit{layout} cues. For example, when items are arranged in a grid, the observation exposes them as a flat list without indicating column-level layout information.
This omission simplifies the representation but removes structural signals that humans rely on to reason about grouping, alignment, and spatial organization.

Second, this work primarily focuses on the design of a scalable and robust RL environment for Web Agents, rather than on improving RL algorithm performance itself. We leave the design of more effective RL algorithms tailored to web agent scenarios as future work.

Finally, we do not conduct large-scale experiments on real, production websites beyond lab-controlled environments.
This limitation is due to company legal constraints, which prevent automated interaction with live commercial platforms at scale because of Terms of Service restrictions.
We fully acknowledge the importance of real-world evaluation and note that this limitation of existing benchmarks is a key motivation behind the design of \projectname.
Unlike prior environments that rely on per-site configuration or site-specific signals such as accessibility trees, \projectname is built with generalization as a core design goal and requires no site-specific customization.
Through its browser-level execution and parsing pipeline, it can robustly handle real-world behaviors such as pop-ups, loading delays, dynamic content, and timeouts.
We hope future work, with proper legal approval, can further investigate the generalizability of agents trained in \projectname to real-world websites under realistic deployment conditions.

\bibliography{custom}

\appendix

\section{Observation Space Details}
\label{sec:observation-space}

This appendix provides full details of the observation construction pipeline in \projectname.

\paragraph{Filtering.} The parser excludes entire classes of tags that are not directly useful for task execution, including \code{<script>}, \code{<style>}, etc. In addition, elements that are invisible due to CSS properties, that render with zero width and height, or that are completely off-screen and non-scrollable are removed. Only a restricted set of safe attributes is preserved, consisting of a whitelist of HTML attributes (e.g., \code{id}, \code{name}, ...).

\paragraph{Flattening and pruning.} To reduce structural redundancy, the parser collapses trivial nesting of non-semantic containers, such as chains of \code{<div>} wrappers, and prunes empty elements. Exceptions are made for controls that may legitimately appear empty (e.g., \code{<input>}, \code{<select>}). Inline text nodes containing non-whitespace characters are preserved in order to retain meaningful labels.

\paragraph{Interactivity detection.} The parser heuristically determines whether elements are interactive. Nodes are marked as clickable if they are native controls (\code{<button>}), anchors with \code{href}, elements with explicit \code{onclick} handlers, or elements with ARIA roles (\code{button}, \code{link}). In addition, elements with computed cursor style \code{pointer} are considered clickable. Elements that are disabled are excluded. To identify hover-sensitive targets, we monkey-patch \code{addEventListener} such that whenever a node registers a hover event listener, the element is annotated with \code{data-maybe-hoverable=true}.

\paragraph{Semantic identifiers.} Each interactive element is assigned a stable, human-readable semantic identifier. Base names are derived from visible text, placeholders, or tag names, normalized to short identifiers. Identifiers are scoped hierarchically by their parent elements' names, and uniqueness is enforced globally by appending numeric suffixes as needed.

\paragraph{Javascript state capture.} The parser augments the stripped DOM with fine-grained state information. For text inputs, textareas, and contenteditable regions, it records the current value, whether the control can be edited, numeric values (for \code{type=number}), text selection ranges, and focus state. For select elements, it records the current value, the selected index, whether multiple selection is enabled, and the set of selected options. Each option is cloned with its text and value, marked as selected when appropriate, and assigned its own semantic identifier.

\paragraph{Output schema.} The final observation is returned as a JSON object with five components: (1) a stripped and annotated HTML snapshot, (2) a list of clickable elements identified by semantic ID, (3) a list of hoverable elements, (4) a list of input elements with their state (identifier, type, value, editability, focus), and (5) a list of select elements with their selection state and per-option identifiers. This representation removes hidden or redundant markup while retaining the cues that humans rely on for interactivity, yielding a compact, semantically aligned observation space.

\section{Action Space}
\label{sec:action-space}
\begin{itemize}[leftmargin=*]

\item {\textbf{A. Element-level interactions}}
  \begin{itemize}[leftmargin=*]
    \item {Click element}: perform a click action on the target element such as a button, link, or control.  
    \item {Hover element}: move the cursor over the target element to trigger tooltips or dropdown menus.  
    \item {Key press}: send a keyboard event such as Enter, Escape, Tab, or arrow keys, optionally focusing an element first.  
  \end{itemize}

\item {\textbf{B. Form and text input}}
  \begin{itemize}[leftmargin=*]
    \item {Type text}: enter text into an input field or editable region, optionally pressing Enter afterwards.  
    \item {Clear input}: remove all content from an input field or editable region.  
    \item {Select option}: choose a specific option from a dropdown or select menu.  
  \end{itemize}

\item {\textbf{C. Navigation and page control}}
  \begin{itemize}[leftmargin=*]
    \item {Navigate to URL}: load a new address in the current tab.  
    \item {Back}: navigate one step backward in the browser history.  
    \item {Forward}: navigate one step forward in the browser history.  
    \item {Refresh}: reload the current page.  
  \end{itemize}

\item {\textbf{D. Tab management and task control}}
  \begin{itemize}[leftmargin=*]
    \item {New tab}: open a new browser tab, optionally with a specified URL.  
    \item {Switch tab}: change focus to another tab by index.  
    \item {Close tab}: close an existing browser tab.  
    \item {Terminate task}: shut down the browser session and optionally submit a final answer.
  \end{itemize}

\end{itemize}

\section{RL Training Hyperparameters}
\label{sec:rl-hyperparams}

We train with GRPO using the following configuration. The advantage estimator is GRPO with dynamic sampling filtering (reward nonzero std). We use low-variance KL loss with coefficient 0.001, entropy coefficient 0.0, and PPO clipping with $\epsilon = 0.2$ (high clip 0.28). The optimizer is Adam with learning rate $1 \times 10^{-6}$ (constant schedule), weight decay 0.01, $\beta_1 = 0.9$, and $\beta_2 = 0.98$. For rollouts, we use a batch size of 16, training batch size of 12, 12 samples per prompt, over-sampling batch size of 32, max response length of 4096 tokens, and temperature 1.0. Training is distributed across 64$\times$ NVIDIA H200 GPUs (4 actor nodes and 4 rollout nodes, each with 8 GPUs).

\section*{NeurIPS Paper Checklist}

The checklist is designed to encourage best practices for responsible machine learning research, addressing issues of reproducibility, transparency, research ethics, and societal impact. Do not remove the checklist: {\bf The papers not including the checklist will be desk rejected.} The checklist should follow the references and follow the (optional) supplemental material.  The checklist does NOT count towards the page
limit. 

Please read the checklist guidelines carefully for information on how to answer these questions. For each question in the checklist:
\begin{itemize}
    \item You should answer \answerYes{}, \answerNo{}, or \answerNA{}.
    \item \answerNA{} means either that the question is Not Applicable for that particular paper or the relevant information is Not Available.
    \item Please provide a short (1--2 sentence) justification right after your answer (even for \answerNA). 
\end{itemize}

{\bf The checklist answers are an integral part of your paper submission.} They are visible to the reviewers, area chairs, senior area chairs, and ethics reviewers. You will also be asked to include it (after eventual revisions) with the final version of your paper, and its final version will be published with the paper.

The reviewers of your paper will be asked to use the checklist as one of the factors in their evaluation. While \answerYes{} is generally preferable to \answerNo{}, it is perfectly acceptable to answer \answerNo{} provided a proper justification is given (e.g., error bars are not reported because it would be too computationally expensive'' or ``we were unable to find the license for the dataset we used''). In general, answering \answerNo{} or \answerNA{} is not grounds for rejection. While the questions are phrased in a binary way, we acknowledge that the true answer is often more nuanced, so please just use your best judgment and write a justification to elaborate. All supporting evidence can appear either in the main paper or the supplemental material, provided in appendix. If you answer \answerYes{} to a question, in the justification please point to the section(s) where related material for the question can be found.

IMPORTANT, please:
\begin{itemize}
    \item {\bf Delete this instruction block, but keep the section heading ``NeurIPS Paper Checklist"},
    \item  {\bf Keep the checklist subsection headings, questions/answers and guidelines below.}
    \item {\bf Do not modify the questions and only use the provided macros for your answers}.
\end{itemize}

\begin{enumerate}

\item {\bf Claims}
    \item[] Question: Do the main claims made in the abstract and introduction accurately reflect the paper's contributions and scope?
    \item[] Answer: \answerYes{}
    \item[] Justification: The abstract and introduction clearly state our three contributions (browser I/O interface, robust action execution, scalable Incus-based server manager) and the experimental results that support them (Tables 1--5).
    \item[] Guidelines:
    \begin{itemize}
        \item The answer \answerNA{} means that the abstract and introduction do not include the claims made in the paper.
        \item The abstract and/or introduction should clearly state the claims made, including the contributions made in the paper and important assumptions and limitations. A \answerNo{} or \answerNA{} answer to this question will not be perceived well by the reviewers. 
        \item The claims made should match theoretical and experimental results, and reflect how much the results can be expected to generalize to other settings. 
        \item It is fine to include aspirational goals as motivation as long as it is clear that these goals are not attained by the paper. 
    \end{itemize}

\item {\bf Limitations}
    \item[] Question: Does the paper discuss the limitations of the work performed by the authors?
    \item[] Answer: \answerYes{}
    \item[] Justification: See the dedicated Limitations section. We discuss lack of visual layout cues in text-only mode, focus on environment rather than RL algorithm design, and inability to evaluate on live production websites due to legal constraints.
    \item[] Guidelines:
    \begin{itemize}
        \item The answer \answerNA{} means that the paper has no limitation while the answer \answerNo{} means that the paper has limitations, but those are not discussed in the paper. 
        \item The authors are encouraged to create a separate ``Limitations'' section in their paper.
        \item The paper should point out any strong assumptions and how robust the results are to violations of these assumptions (e.g., independence assumptions, noiseless settings, model well-specification, asymptotic approximations only holding locally). The authors should reflect on how these assumptions might be violated in practice and what the implications would be.
        \item The authors should reflect on the scope of the claims made, e.g., if the approach was only tested on a few datasets or with a few runs. In general, empirical results often depend on implicit assumptions, which should be articulated.
        \item The authors should reflect on the factors that influence the performance of the approach. For example, a facial recognition algorithm may perform poorly when image resolution is low or images are taken in low lighting. Or a speech-to-text system might not be used reliably to provide closed captions for online lectures because it fails to handle technical jargon.
        \item The authors should discuss the computational efficiency of the proposed algorithms and how they scale with dataset size.
        \item If applicable, the authors should discuss possible limitations of their approach to address problems of privacy and fairness.
        \item While the authors might fear that complete honesty about limitations might be used by reviewers as grounds for rejection, a worse outcome might be that reviewers discover limitations that aren't acknowledged in the paper. The authors should use their best judgment and recognize that individual actions in favor of transparency play an important role in developing norms that preserve the integrity of the community. Reviewers will be specifically instructed to not penalize honesty concerning limitations.
    \end{itemize}

\item {\bf Theory assumptions and proofs}
    \item[] Question: For each theoretical result, does the paper provide the full set of assumptions and a complete (and correct) proof?
    \item[] Answer: \answerNA{}
    \item[] Justification: This paper does not include theoretical results.
    \item[] Guidelines:
    \begin{itemize}
        \item The answer \answerNA{} means that the paper does not include theoretical results. 
        \item All the theorems, formulas, and proofs in the paper should be numbered and cross-referenced.
        \item All assumptions should be clearly stated or referenced in the statement of any theorems.
        \item The proofs can either appear in the main paper or the supplemental material, but if they appear in the supplemental material, the authors are encouraged to provide a short proof sketch to provide intuition. 
        \item Inversely, any informal proof provided in the core of the paper should be complemented by formal proofs provided in appendix or supplemental material.
        \item Theorems and Lemmas that the proof relies upon should be properly referenced. 
    \end{itemize}

    \item {\bf Experimental result reproducibility}
    \item[] Question: Does the paper fully disclose all the information needed to reproduce the main experimental results of the paper to the extent that it affects the main claims and/or conclusions of the paper (regardless of whether the code and data are provided or not)?
    \item[] Answer: \answerYes{}
    \item[] Justification: We describe the full environment design (Section 3), evaluation protocol (Section 4), system configuration (AWS instance type, container settings), and RL training hyperparameters (Appendix C). Code is available at \url{https://anonymous.4open.science/r/webserv_anonymous-90EF/}.
    \item[] Guidelines:
    \begin{itemize}
        \item The answer \answerNA{} means that the paper does not include experiments.
        \item If the paper includes experiments, a \answerNo{} answer to this question will not be perceived well by the reviewers: Making the paper reproducible is important, regardless of whether the code and data are provided or not.
        \item If the contribution is a dataset and\slash or model, the authors should describe the steps taken to make their results reproducible or verifiable. 
        \item Depending on the contribution, reproducibility can be accomplished in various ways. For example, if the contribution is a novel architecture, describing the architecture fully might suffice, or if the contribution is a specific model and empirical evaluation, it may be necessary to either make it possible for others to replicate the model with the same dataset, or provide access to the model. In general. releasing code and data is often one good way to accomplish this, but reproducibility can also be provided via detailed instructions for how to replicate the results, access to a hosted model (e.g., in the case of a large language model), releasing of a model checkpoint, or other means that are appropriate to the research performed.
        \item While NeurIPS does not require releasing code, the conference does require all submissions to provide some reasonable avenue for reproducibility, which may depend on the nature of the contribution. For example
        \begin{enumerate}
            \item If the contribution is primarily a new algorithm, the paper should make it clear how to reproduce that algorithm.
            \item If the contribution is primarily a new model architecture, the paper should describe the architecture clearly and fully.
            \item If the contribution is a new model (e.g., a large language model), then there should either be a way to access this model for reproducing the results or a way to reproduce the model (e.g., with an open-source dataset or instructions for how to construct the dataset).
            \item We recognize that reproducibility may be tricky in some cases, in which case authors are welcome to describe the particular way they provide for reproducibility. In the case of closed-source models, it may be that access to the model is limited in some way (e.g., to registered users), but it should be possible for other researchers to have some path to reproducing or verifying the results.
        \end{enumerate}
    \end{itemize}

\item {\bf Open access to data and code}
    \item[] Question: Does the paper provide open access to the data and code, with sufficient instructions to faithfully reproduce the main experimental results, as described in supplemental material?
    \item[] Answer: \answerYes{}
    \item[] Justification: Code and data are available at \url{https://anonymous.4open.science/r/webserv_anonymous-90EF/}. The evaluation tasks are from the publicly available WebArena-Lite benchmark.
    \item[] Guidelines:
    \begin{itemize}
        \item The answer \answerNA{} means that paper does not include experiments requiring code.
        \item Please see the NeurIPS code and data submission guidelines (\url{https://neurips.cc/public/guides/CodeSubmissionPolicy}) for more details.
        \item While we encourage the release of code and data, we understand that this might not be possible, so \answerNo{} is an acceptable answer. Papers cannot be rejected simply for not including code, unless this is central to the contribution (e.g., for a new open-source benchmark).
        \item The instructions should contain the exact command and environment needed to run to reproduce the results. See the NeurIPS code and data submission guidelines (\url{https://neurips.cc/public/guides/CodeSubmissionPolicy}) for more details.
        \item The authors should provide instructions on data access and preparation, including how to access the raw data, preprocessed data, intermediate data, and generated data, etc.
        \item The authors should provide scripts to reproduce all experimental results for the new proposed method and baselines. If only a subset of experiments are reproducible, they should state which ones are omitted from the script and why.
        \item At submission time, to preserve anonymity, the authors should release anonymized versions (if applicable).
        \item Providing as much information as possible in supplemental material (appended to the paper) is recommended, but including URLs to data and code is permitted.
    \end{itemize}

\item {\bf Experimental setting/details}
    \item[] Question: Does the paper specify all the training and test details (e.g., data splits, hyperparameters, how they were chosen, type of optimizer) necessary to understand the results?
    \item[] Answer: \answerYes{}
    \item[] Justification: Training hyperparameters are provided in Appendix C. System benchmarks specify the hardware (AWS EC2 r6id.metal, 128 vCPUs, 1024 GiB memory). RL training specifies GPU configuration (64$\times$ H200). Evaluation uses the standard WebArena-Lite 110-task protocol.
    \item[] Guidelines:
    \begin{itemize}
        \item The answer \answerNA{} means that the paper does not include experiments.
        \item The experimental setting should be presented in the core of the paper to a level of detail that is necessary to appreciate the results and make sense of them.
        \item The full details can be provided either with the code, in appendix, or as supplemental material.
    \end{itemize}

\item {\bf Experiment statistical significance}
    \item[] Question: Does the paper report error bars suitably and correctly defined or other appropriate information about the statistical significance of the experiments?
    \item[] Answer: \answerNo{}
    \item[] Justification: We report single-run results following the standard WebArena-Lite evaluation protocol used by prior work. Error bars are not reported due to the high computational cost of repeated full evaluations across multiple models.
    \item[] Guidelines:
    \begin{itemize}
        \item The answer \answerNA{} means that the paper does not include experiments.
        \item The authors should answer \answerYes{} if the results are accompanied by error bars, confidence intervals, or statistical significance tests, at least for the experiments that support the main claims of the paper.
        \item The factors of variability that the error bars are capturing should be clearly stated (for example, train/test split, initialization, random drawing of some parameter, or overall run with given experimental conditions).
        \item The method for calculating the error bars should be explained (closed form formula, call to a library function, bootstrap, etc.)
        \item The assumptions made should be given (e.g., Normally distributed errors).
        \item It should be clear whether the error bar is the standard deviation or the standard error of the mean.
        \item It is OK to report 1-sigma error bars, but one should state it. The authors should preferably report a 2-sigma error bar than state that they have a 96\% CI, if the hypothesis of Normality of errors is not verified.
        \item For asymmetric distributions, the authors should be careful not to show in tables or figures symmetric error bars that would yield results that are out of range (e.g., negative error rates).
        \item If error bars are reported in tables or plots, the authors should explain in the text how they were calculated and reference the corresponding figures or tables in the text.
    \end{itemize}

\item {\bf Experiments compute resources}
    \item[] Question: For each experiment, does the paper provide sufficient information on the computer resources (type of compute workers, memory, time of execution) needed to reproduce the experiments?
    \item[] Answer: \answerYes{}
    \item[] Justification: System benchmarks use an AWS EC2 r6id.metal instance (128 vCPUs, 1024 GiB). RL training uses 64$\times$ NVIDIA H200 GPUs across 8 nodes, with 4 CPU instances for container hosting. Each RL step takes approximately 12 minutes (Section 4.3.1).
    \item[] Guidelines:
    \begin{itemize}
        \item The answer \answerNA{} means that the paper does not include experiments.
        \item The paper should indicate the type of compute workers CPU or GPU, internal cluster, or cloud provider, including relevant memory and storage.
        \item The paper should provide the amount of compute required for each of the individual experimental runs as well as estimate the total compute. 
        \item The paper should disclose whether the full research project required more compute than the experiments reported in the paper (e.g., preliminary or failed experiments that didn't make it into the paper). 
    \end{itemize}
    
\item {\bf Code of ethics}
    \item[] Question: Does the research conducted in the paper conform, in every respect, with the NeurIPS Code of Ethics \url{https://neurips.cc/public/EthicsGuidelines}?
    \item[] Answer: \answerYes{}
    \item[] Justification: The research conforms with the NeurIPS Code of Ethics. We do not interact with live production websites at scale, and our environment operates on self-hosted Docker images.
    \item[] Guidelines:
    \begin{itemize}
        \item The answer \answerNA{} means that the authors have not reviewed the NeurIPS Code of Ethics.
        \item If the authors answer \answerNo, they should explain the special circumstances that require a deviation from the Code of Ethics.
        \item The authors should make sure to preserve anonymity (e.g., if there is a special consideration due to laws or regulations in their jurisdiction).
    \end{itemize}

\item {\bf Broader impacts}
    \item[] Question: Does the paper discuss both potential positive societal impacts and negative societal impacts of the work performed?
    \item[] Answer: \answerNA{}
    \item[] Justification: This work provides infrastructure for training web agents in sandboxed environments. It does not directly enable harmful applications beyond what existing web automation tools already provide.
    \item[] Guidelines:
    \begin{itemize}
        \item The answer \answerNA{} means that there is no societal impact of the work performed.
        \item If the authors answer \answerNA{} or \answerNo, they should explain why their work has no societal impact or why the paper does not address societal impact.
        \item Examples of negative societal impacts include potential malicious or unintended uses (e.g., disinformation, generating fake profiles, surveillance), fairness considerations (e.g., deployment of technologies that could make decisions that unfairly impact specific groups), privacy considerations, and security considerations.
        \item The conference expects that many papers will be foundational research and not tied to particular applications, let alone deployments. However, if there is a direct path to any negative applications, the authors should point it out. For example, it is legitimate to point out that an improvement in the quality of generative models could be used to generate Deepfakes for disinformation. On the other hand, it is not needed to point out that a generic algorithm for optimizing neural networks could enable people to train models that generate Deepfakes faster.
        \item The authors should consider possible harms that could arise when the technology is being used as intended and functioning correctly, harms that could arise when the technology is being used as intended but gives incorrect results, and harms following from (intentional or unintentional) misuse of the technology.
        \item If there are negative societal impacts, the authors could also discuss possible mitigation strategies (e.g., gated release of models, providing defenses in addition to attacks, mechanisms for monitoring misuse, mechanisms to monitor how a system learns from feedback over time, improving the efficiency and accessibility of ML).
    \end{itemize}
    
\item {\bf Safeguards}
    \item[] Question: Does the paper describe safeguards that have been put in place for responsible release of data or models that have a high risk for misuse (e.g., pre-trained language models, image generators, or scraped datasets)?
    \item[] Answer: \answerNA{}
    \item[] Justification: The paper releases an environment infrastructure tool, not a pre-trained model or scraped dataset. The environment operates on self-hosted containers with no access to external services.
    \item[] Guidelines:
    \begin{itemize}
        \item The answer \answerNA{} means that the paper poses no such risks.
        \item Released models that have a high risk for misuse or dual-use should be released with necessary safeguards to allow for controlled use of the model, for example by requiring that users adhere to usage guidelines or restrictions to access the model or implementing safety filters. 
        \item Datasets that have been scraped from the Internet could pose safety risks. The authors should describe how they avoided releasing unsafe images.
        \item We recognize that providing effective safeguards is challenging, and many papers do not require this, but we encourage authors to take this into account and make a best faith effort.
    \end{itemize}

\item {\bf Licenses for existing assets}
    \item[] Question: Are the creators or original owners of assets (e.g., code, data, models), used in the paper, properly credited and are the license and terms of use explicitly mentioned and properly respected?
    \item[] Answer: \answerYes{}
    \item[] Justification: We cite WebArena and WebArena-Lite, which are publicly available under open-source licenses. We use Incus (Apache 2.0) and standard open-source web frameworks.
    \item[] Guidelines:
    \begin{itemize}
        \item The answer \answerNA{} means that the paper does not use existing assets.
        \item The authors should cite the original paper that produced the code package or dataset.
        \item The authors should state which version of the asset is used and, if possible, include a URL.
        \item The name of the license (e.g., CC-BY 4.0) should be included for each asset.
        \item For scraped data from a particular source (e.g., website), the copyright and terms of service of that source should be provided.
        \item If assets are released, the license, copyright information, and terms of use in the package should be provided. For popular datasets, \url{paperswithcode.com/datasets} has curated licenses for some datasets. Their licensing guide can help determine the license of a dataset.
        \item For existing datasets that are re-packaged, both the original license and the license of the derived asset (if it has changed) should be provided.
        \item If this information is not available online, the authors are encouraged to reach out to the asset's creators.
    \end{itemize}

\item {\bf New assets}
    \item[] Question: Are new assets introduced in the paper well documented and is the documentation provided alongside the assets?
    \item[] Answer: \answerYes{}
    \item[] Justification: WebServ code with documentation is available at \url{https://anonymous.4open.science/r/webserv_anonymous-90EF/}. The code includes installation instructions, usage examples, and integration with VeRL/Slime for RL training.
    \item[] Guidelines:
    \begin{itemize}
        \item The answer \answerNA{} means that the paper does not release new assets.
        \item Researchers should communicate the details of the dataset\slash code\slash model as part of their submissions via structured templates. This includes details about training, license, limitations, etc. 
        \item The paper should discuss whether and how consent was obtained from people whose asset is used.
        \item At submission time, remember to anonymize your assets (if applicable). You can either create an anonymized URL or include an anonymized zip file.
    \end{itemize}

\item {\bf Crowdsourcing and research with human subjects}
    \item[] Question: For crowdsourcing experiments and research with human subjects, does the paper include the full text of instructions given to participants and screenshots, if applicable, as well as details about compensation (if any)? 
    \item[] Answer: \answerNA{}
    \item[] Justification: This paper does not involve crowdsourcing or research with human subjects.
    \item[] Guidelines:
    \begin{itemize}
        \item The answer \answerNA{} means that the paper does not involve crowdsourcing nor research with human subjects.
        \item Including this information in the supplemental material is fine, but if the main contribution of the paper involves human subjects, then as much detail as possible should be included in the main paper. 
        \item According to the NeurIPS Code of Ethics, workers involved in data collection, curation, or other labor should be paid at least the minimum wage in the country of the data collector. 
    \end{itemize}

\item {\bf Institutional review board (IRB) approvals or equivalent for research with human subjects}
    \item[] Question: Does the paper describe potential risks incurred by study participants, whether such risks were disclosed to the subjects, and whether Institutional Review Board (IRB) approvals (or an equivalent approval/review based on the requirements of your country or institution) were obtained?
    \item[] Answer: \answerNA{}
    \item[] Justification: This paper does not involve research with human subjects.
    \item[] Guidelines:
    \begin{itemize}
        \item The answer \answerNA{} means that the paper does not involve crowdsourcing nor research with human subjects.
        \item Depending on the country in which research is conducted, IRB approval (or equivalent) may be required for any human subjects research. If you obtained IRB approval, you should clearly state this in the paper. 
        \item We recognize that the procedures for this may vary significantly between institutions and locations, and we expect authors to adhere to the NeurIPS Code of Ethics and the guidelines for their institution. 
        \item For initial submissions, do not include any information that would break anonymity (if applicable), such as the institution conducting the review.
    \end{itemize}

\item {\bf Declaration of LLM usage}
    \item[] Question: Does the paper describe the usage of LLMs if it is an important, original, or non-standard component of the core methods in this research? Note that if the LLM is used only for writing, editing, or formatting purposes and does \emph{not} impact the core methodology, scientific rigor, or originality of the research, declaration is not required.
    \item[] Answer: \answerNA{}
    \item[] Justification: LLM code writing tools were used to assist in writing framework code, which does not impact the core methodology or originality of the research.
    \item[] Guidelines:
    \begin{itemize}
        \item The answer \answerNA{} means that the core method development in this research does not involve LLMs as any important, original, or non-standard components.
        \item Please refer to our LLM policy in the NeurIPS handbook for what should or should not be described.
    \end{itemize}

\end{enumerate}
\end{document}